\pdfoutput=1
\documentclass{article} 
\usepackage{arxiv}  
\usepackage{times}  
\usepackage{helvet} 
\usepackage{courier}  
\usepackage[hyphens]{url}  
\usepackage{graphicx} 
\usepackage{graphicx}  
\usepackage{amsmath,amsopn,amssymb}
\usepackage{algorithm}
\usepackage[noend]{algpseudocode}
\usepackage{comment}
\usepackage{subfig}
\usepackage{multirow}

\title{Joint Graph and Feature Learning In Graph Convolutional Neural Networks }

\author{Jiaxiang Tang, Wei Hu, Xiang Gao, Zongming Guo \\
Institute of Computer Science and Technology, Peking University, China \\
\small \{hawkey1999, forhuwei, gyshgx868, guozongming\}@pku.edu.cn}

\begin{document}

\maketitle

\begin{abstract}
Graph Convolutional Neural Networks (GCNNs) extend classical CNNs to graph data domain, such as brain networks, social networks and 3D point clouds. 
It is critical to identify an appropriate graph for the subsequent graph convolution. 
Existing methods manually construct or learn one fixed graph for all the layers of a GCNN. 
In order to adapt to the underlying structure of node features in different layers, we propose dynamic learning of graphs and node features jointly in GCNNs. 
In particular, we cast the graph optimization problem as distance metric learning to capture pairwise similarities of features in each layer. 
We deploy the Mahalanobis distance metric and further decompose the metric matrix into a low-dimensional matrix, which converts graph learning to the optimization of a low-dimensional matrix for efficient implementation. 
Extensive experiments on point clouds and citation network datasets demonstrate the superiority of the proposed method in terms of both accuracies and robustness. 
\end{abstract}

\section{Introduction}
\label{sec:intro}
Graph Convolutional Neural Networks (GCNNs) have been receiving increasing attention as a powerful tool for irregularly structured data on graphs, such as citation networks, social networks and 3D point clouds.  
The construction of an appropriate graph topology plays a critical role in GCNNs for efficient feature learning. 
In settings where the graph is inaccurate or even not readily available, it is necessary to infer or learn a graph topology from data before it is used for guiding graph convolution in GCNNs.   

Most of the previous studies construct underlying graphs from data empirically, such as $k$-Nearest-Neighbor ($k$-NN) graphs \cite{wang2019dgcnn,te2018rgcnn}, which may lead to sub-optimal solutions.
Few methods exploit \textit{graph learning} for optimized representation \cite{Jiang2018GraphLN,li2018adaptive,shi2018non,li2019spatio}, which learns a fixed and shared graph for all instances, or an individual graph for each instance, or a combination of shared and individual graphs.  
However, only \textit{one fixed} graph is learned and applied to all layers of the entire network, which may not well capture the underlying structure of node features in different layers dynamically. 

Extending on these previous studies, we propose a Joint Learning Graph Convolutional Network (JLGCN), which exploits \textit{dynamic} learning of graphs and node features \textit{jointly} in GCNNs. 
In particular, we optimize an underlying graph kernel from data features via distance metric learning that characterizes pairwise similarities of data.  
We deploy the \textit{Mahalanobis distance} \cite{mahalanobis1936}, which takes into account data correlations for intrinsic representation. 
Given a K-dimensional feature vector $\mathbf{f}_i$ per node $i$, $\mathbf{f}_i \in \mathbb{R}^K$, the Mahalanobis distance between the features $\mathbf{f}_i$ and $\mathbf{f}_j$ of nodes $i$ and $j$ is defined as:
$d_{\mathbf{M}}(\mathbf{f}_i,\mathbf{f}_j)=\sqrt{(\mathbf{f}_i - \mathbf{f}_j)^{\top} \mathbf{M} (\mathbf{f}_i - \mathbf{f}_j)}$, where $\mathbf{M} \in \mathbb{R}^{K \times K}$ is the Mahalanobis distance metric matrix which reflects feature correlations. 
Hence, we convert the problem of graph learning to the optimization of $\mathbf{M}$. 
As $\mathbf{M}$ is positive semi-definite (PSD), it is often nontrivial to solve efficiently. 
Instead, we decompose it as $\mathbf{M} = \mathbf{R} \mathbf{R}^{\top}$ and learn $\mathbf{R}$ for ease of optimization, where $\mathbf{R} \in \mathbb{R}^{K \times S}$ has a lower dimension $S<<K$ to reduce number of parameters for efficient implementation. 
Given features $\mathbf{f}$, we seek to minimize the Graph Laplacian Regularizer (GLR) \cite{shuman2013emerging} $\mathbf{f}^{\top}\mathbf{L}(\mathbf{R})\mathbf{f}$ by optimizing $\mathbf{R}$, which measures the smoothness of features $\mathbf{f}$ with respect to the graph Laplacian $\mathbf{L}$\footnote{In spectral graph theory, a graph Laplacian matrix is an algebraic representation of connectivities and node degrees of the corresponding graph, which will be defined formally later.}.  
This essentially enforces the graph encoded in $\mathbf{L}(\mathbf{R})$ to capture pairwise similarities of $\mathbf{f}$.  

Hence, we formulate the joint learning of graphs $\mathbf{L}(\mathbf{R})$ and node features $\mathbf{f}$ as an optimization problem, which minimizes a weighted sum of the GLR and cross-entropy. 
We set this objective as the loss function of the proposed JLGCN to guide network model optimization, which employs a localized first-order approximation of spectral graph convolution as in \cite{Kipf2016SemiSupervisedCW}.
Further, the learned graph at the previous layer is added to that of the current layer for multi-level feature learning. 
The proposed JLGCN can be integrated into any GCNN architecture for applications such as node classification and graph classification. 
To validate the effectiveness of the proposed JLGCN, we apply it to semi-supervised learning for citation networks and 3D point cloud learning problems.
Extensive experimental results demonstrate the superiority and robustness of JLGCN compared with state-of-the-art methods on four datasets even with a small model size.  

Our contributions can be summarized as follows:
\begin{itemize}
\item We propose joint learning of underlying graphs and node features at each layer of a GCNN, which captures pairwise similarities of node features dynamically. 
\item We cast the graph learning problem as distance metric learning with the Mahalanobis distance deployed. We further decompose the distance metric into a low-dimensional matrix for efficient implementation, which is optimized from both the GLR and cross entropy along with node features. 
\item Extensive experiments on semi-supervised learning and point cloud classification demonstrate the superiority and robustness of our method. 
\end{itemize}

\section{Related Work}
\label{sec:related}
\subsection{Graph Convolution Neural Networks}
As a generalization of CNNs to irregular graph domain, GCNNs can be categorized into two main classes: spectral methods and spatial methods. 

\subsubsection{Spectral methods} This class of methods define graph convolution based on the spectral representation of graphs. 
\cite{bruna2013spectral} defines the convolution in graph Fourier transform domain by eigen-decomposition of the graph Laplacian, but requires intense computations. 
\cite{defferrard2016convolutional} addresses this problem by leveraging Chebyshev polynomials to approximate spectral filters and achieves localized filtering. 
GCN \cite{Kipf2016SemiSupervisedCW} further simplifies the previous work by employing only first-order approximation of the filters.
Our proposed joint learning of graphs and node features is based on GCN, which has spectral interpretation and scales linearly with the number of edges.

\subsubsection{Spatial methods} Instead of spectral representation, spatial methods define graph convolution directly on each node and its neighbors for feature propagation. 
Mixture model network \cite{monti2017geometric} provides a unified generalization of CNN architectures on graphs. Graph attention network (GAT) in \cite{velivckovic2017graph} employs self-attention mechanism to solve the node classification problem on citation networks. DGCNN \cite{wang2019dgcnn} proposes edge convolution to aggregate local features, which is applied to point cloud learning problem.

\subsection{Graph Learning in GCNNs}
Graph Learning in GCNNs can be categorized into three classes based on the graph domain to address: 
1) fixed-domain graph learning, in which the underlying graph is fixed; 
2) varying-domain graph learning, where graphs vary across data instances; 
and 3) hybrid graph learning, which combines the former two classes. 

\subsubsection{Fixed-domain Graph Learning}
This class of graph learning assumes there is a shared graph structure underlying all instances with fixed vertices. 
In applications such as semi-supervised node classification of citation networks, there is only \textit{one fixed} graph to be learned and shared by all the instances to classify. 
GLCN \cite{Jiang2018GraphLN} learns a non-negative function that represents the pairwise relationship between two vertices using attention mechanism via a single-layer neural network, and optimizes it by minimizing GLR in the loss function. Instead, our method models pairwise similarity more explicitly by adopting Mahalanobis distance, and we learn different graphs at different layers.

\subsubsection{Varying-domain Graph Learning}
Instances corresponding to varying domains may require different graphs with possibly arbitrary number of vertices. 
Graph learning for such data applications learns adaptive graphs tailored for different individuals. 
Spatio-temporal graph routing (STGR) is proposed \cite{li2019spatio} to learn one spatial graph and one temporal graph for each skeleton instance.
Adaptive graph convolution network (AGCN) is proposed \cite{li2018adaptive} to construct a unique residual Laplacian matrix by learning Mahalanobis distance metric for each instance.
Different from the task-driven AGCN where the Mahalanobis distance metric $\mathbf{M}$ is learned by minimizing the cross-entropy only, we optimize $\mathbf{M}$ via both the GLR and cross-entropy.
Also, we assume $\mathbf{M}$ is low rank and set $S << K$ to reduce number of parameters while AGCN adopts $S = K$; 

\subsubsection{Hybrid Graph Learning}
For special applications like skeleton-based action recognition, both fixed-domain and varying-domain graph learning can be applied according to the way of graph construction.
NLGCN \cite{shi2018non} builds three graphs for each instance: the pre-defined physical skeleton graph, a shared graph for all instances, and an individual graph learned from each instance's own feature in a non-local \cite{wang2018non} manner.

\vspace{-0.05in}
\section{Background in Spectral Graph Theory}
\label{sec:graph}
\subsection{Graph and Graph Laplacian}
An undirected graph $\mathcal{G}=\{\mathcal{V},\mathcal{E},\mathbf{A}\}$ is composed of a vertex set $\mathcal{V}$ of cardinality $\left|\mathcal{V}\right|=N$, an edge set $\mathcal{E}$ connecting vertices, and a weighted adjacency matrix $\mathbf{A}$. $\mathbf{A} \in \mathbb{R}^{N \times N}$ is a real and symmetric matrix, where $a_{i,j} \geq 0$ is the weight assigned to the edge $(i,j)$ connecting vertices $i$ and $j$. 
Edge weights often measure the similarity between connected vertices. 

The graph Laplacian matrix is defined from the adjacency matrix. 
Among different variants of Laplacian matrices, the \textit{combinatorial graph Laplacian} is defined as $\mathbf{L}:=\mathbf{D}-\mathbf{A}$, where $\mathbf{D}$ is the \textit{degree matrix}---a diagonal matrix where $d_{i,i}=\sum_{j=1}^N a_{i,j}$.

\subsection{Graph Laplacian Regularizer} 
Graph signal refers to data that resides on the nodes of a graph, such as temperatures on a sensor network. A graph signal $\mathbf{x}\in\mathbb{R}^N$ defined on a graph $\mathcal{G}$ is \textit{smooth} with respect to $\mathcal{G}$ \cite{spielman2007spectral} if  
\begin{equation}
	\mathbf{x}^{\top} \mathbf{L} \mathbf{x} =\sum_{i=1}^{N} \sum_{j=1}^{N} w_{i,j}(x_i - x_j)^2 < \epsilon,
	\label{eq:prior}
\end{equation}
where $\epsilon$ is a small positive scalar. 
To satisfy Eq. \eqref{eq:prior}, connected node pair $x_i$ and $x_j$ must be similar for a large edge weight $w_{i,j}$; for a small $w_{i,j}$, $x_i$ and $x_j$ can differ significantly. Hence, Eq. \eqref{eq:prior} forces $\mathbf{x}$ to adapt to the topology of $\mathcal{G}$, and is commonly called the \textit{graph Laplacian Regularizer} (GLR) \cite{shuman2013emerging,pang2017graph}.

\vspace{-0.05in}
\section{Joint Learning of Graphs and Features}
\label{sec:method}
In this section, we first propose the problem formulations of graph learning and node feature learning respectively assuming one of the other is known, and then combine them to the formulation of jointly learning.

\subsection{Problem Formulation of Graph Learning}
As a graph essentially captures \textit{pairwise similarities} within data $\mathbf{x} \in \mathbb{R}^N$, we cast the problem of graph learning as \textit{distance metric learning}, \textit{i.e.}, learning a distance function for each data pair $\{x_i,x_j\}$ for similarity calculation. 
Specifically, given K-dimensional feature vectors $\mathbf{f}^i_l \in \mathbb{R}^K$ and $\mathbf{f}^j_l \in \mathbb{R}^K$ of node $i$ and $j$ at the $l$-th layer respectively, we employ the commonly used Gaussian kernel to define an edge weight $a_{i,j}$ as
\begin{equation}
    a_{i,j}=\exp\left\{-d^2(\mathbf{f}^i_l,\mathbf{f}^j_l)\right\},
\end{equation}
where $d(\mathbf{f}^i_l,\mathbf{f}^j_l)$ is a distance metric between $\mathbf{f}^i_l$ and $\mathbf{f}^j_l$. The Gaussian kernel enforces edge weights to be in range $[0,1]$, thus ensuring the resulting combinatorial graph Laplacian to be PSD \cite{cheung2018graph}. 

While there exist various definitions of distance metrics, such as Euclidean distance and bilateral filtering distance \cite{tomasi1998bilateral}, we deploy the \textit{Mahalanobis distance}, which is defined as
\begin{equation}
d_{\mathbf{M}_l}(\mathbf{f}^i_l,\mathbf{f}^j_l) = \sqrt{(\mathbf{f}^i_l - \mathbf{f}^j_l)^{\top} \mathbf{M}_l (\mathbf{f}^i_l - \mathbf{f}^j_l)},
\label{eq:m_distance}
\end{equation}
where $\mathbf{M}_l \in \mathbb{R}^{K \times K}$ is a PSD matrix, and $l$ is the index of layers.
The Mahalanobis distance captures correlations among features via $\mathbf{M}_l$, which is widely adopted in the machine learning literature \cite{li2018adaptive,luo2019robust}. 

Each edge weight is then computed as
\begin{equation}
a_{i,j} = \exp\left\{-(\mathbf{f}^i_l - \mathbf{f}^j_l)^{\top} \mathbf{M}_l (\mathbf{f}^i_l - \mathbf{f}^j_l)\right\},
\label{eq:gaussian_kernel}
\end{equation}
from which we can calculate the corresponding adjacency matrix and graph Laplacian. 

As the graph Laplacian can be computed from edge weights by definition,
we convert the graph learning problem to an optimization problem over $\mathbf{M}_l$. 
We optimize $\mathbf{M}_l$ by minimizing Graph Laplacian Regularizer in Eq. \eqref{eq:prior}, which enforces the graph to capture the underlying structure of features. 
Let $s_{i,j} = (x_i-x_j)^2$, we have the following problem formulation for graph learning over all the layers of the network, with the objective as the sum of GLR over all the layers:
\begin{equation}
\begin{split}
&\begin{split}
	\min_{\mathbf{M}_l}
     & \sum_l \sum_{\{i,j\}}\exp\left\{-(\mathbf{f}^i_l-\mathbf{f}^j_l)^{\top} \mathbf{M}_l (\mathbf{f}^i_l-\mathbf{f}^j_l) \right\} \, s_{i,j}
     \end{split} \\
& \text{s.t.} \quad \,\mathbf{M}_l \succeq 0, \, \forall l.
\label{eq:optimization}
\end{split}
\end{equation}

Since $\mathbf{M}_l$ is PSD and symmetric, Eq. \eqref{eq:optimization} is computationally expensive to solve in general. 
Instead, we decompose $\mathbf{M}_l$ into
\begin{equation}
    \mathbf{M}_l = \mathbf{R}_l \mathbf{R}^{\top}_l,
    \label{eq:learnable_r}
\end{equation}
where $\mathbf{R}_l \in \mathbb{R}^{K \times S}$, and $S \le K$ is a hyper-parameter to control the complexity of computing distance between high-dimensional features.
When $K$ is large and $S << K$, the number of parameters for $\mathbf{M}_l$ is significantly reduced while assuming $\mathbf{M}_l$ is low rank, which is a common assumption in distance metric learning \cite{liu15aaai,luo2019robust}. 

Then the Mahalanobis distance in Eq. \eqref{eq:m_distance} becomes
\begin{equation}
\begin{split}
d_{\mathbf{M}_l}(\mathbf{f}^i_l,\mathbf{f}^j_l)  
&= \sqrt{(\mathbf{f}^i_l - \mathbf{f}^j_l)^{\top} \mathbf{R}_l \mathbf{R}^{\top}_l (\mathbf{f}^i_l - \mathbf{f}^j_l)} \\
&= \sqrt{(\mathbf{R}^{\top}_l (\mathbf{f}^i_l - \mathbf{f}^j_l))^{\top}\mathbf{R}^{\top}_l (\mathbf{f}^i_l - \mathbf{f}^j_l)}\\
&= \|\mathbf{R}^{\top}_l (\mathbf{f}^i_l - \mathbf{f}^j_l)\|_2, 
\end{split}
\label{eq:m_distance_decompose}
\end{equation}
which is essentially a linear transformation of the Euclidean distance by $\mathbf{R}_l$. 
When $\mathbf{R}_l$ is an identity matrix, $d_{\mathbf{M}_l}(\mathbf{f}^i_l,\mathbf{f}^j_l)$ defaults to the Euclidean distance. 

Hence, each edge weight can be computed as 
\begin{equation}
a_{i,j} = \exp\left\{-\|\mathbf{R}^{\top}_l (\mathbf{f}^i_l - \mathbf{f}^j_l)\|_2^2 \right\}.
\label{eq:weight_R}
\end{equation}
Accordingly, the optimization problem in Eq. \eqref{eq:optimization} converts to 
\begin{equation}
    \min_{\{\mathbf{R}_l\}}
     \sum_l \sum_{\{i,j\}}\exp\left\{-\|\mathbf{R}^{\top}_l (\mathbf{f}^i_l - \mathbf{f}^j_l)\|_2^2 \right\} \, s_{i,j},
     \label{eq:opt_graph_R}
\end{equation}
which allows to remove the PSD constraint of $\mathbf{M}_l$ and is thus more efficient to solve.
The optimal adjacency matrix $\mathbf{A}_l^*$ at the $l$-th layer is then computed from the optimized $\mathbf{R}_l$ via Eq. \eqref{eq:weight_R}.  

\subsection{Problem Formulation of Node Feature Learning}
Assuming the availability of the learned optimal graph encoded in $\mathbf{A}_l^*$ and the optimal graph $\mathbf{A}_{l-1}$ in the previous layer, we learn node features based on a modified graph spectral convolution in GCN \cite{Kipf2016SemiSupervisedCW} for its simplicity and effectiveness. 
Given features of all nodes $\mathbf{F}_{l-1}$ from the $(l-1)$-th layer, the graph convolution at the $l$-th layer computes the output feature $\mathbf{F}_{l}$ as:
\begin{equation}
\mathbf{F}_{l} = \mathbf{A}_l\mathbf{F}_{l-1} \mathbf{W}_l
\label{eq:gcn}
\end{equation}
where $\mathbf{W}_l$ is a trainable parameter, and $\mathbf{A}_l$ is the re-normalized adjacency matrix calculated as
\begin{equation}
    \mathbf{A}_l=\mathbf{\Lambda}^{-\frac{1}{2}} (\mathbf{A}_{l-1}+\mathbf{A}_l^*) \mathbf{\Lambda}^{-\frac{1}{2}}
\end{equation}
$\mathbf{\Lambda}$ is the normalization diagonal matrix. 
We essentially replace the re-normalization matrix $\mathbf{I}_N$ in the original GCN layer with our learned optimal graph $\mathbf{A}_l^*$.
When $l=1$, $\mathbf{A}_0=\mathbf{0}$ is a zero matrix.
In special cases where a ground truth graph is provided such as in semi-supervised node classification, $\mathbf{A}_0$ is the ground truth graph. 
The ground truth graph may be inaccurate or could be improved further. 
For instance, in a citation network, an unweighted graph is often available to describe the citation relationship among papers. 
We can further learn a weighted graph to learn hidden connections and characterize how well correlated connected papers are, which enhances the model capacity. 

The objective of node feature learning is to minimize the cross entropy between predicted labels and ground truth labels, which is task-driven. 
We predict the label of each node by applying the softmax function to the output feature at the final $L$-the layer:
\begin{equation}
    \mathbf{\hat Y} = \text{softmax}(\mathbf{F}_L).
\end{equation}
As $\mathbf{F}_L$ is learned from Eq. \eqref{eq:gcn}, $\mathbf{\hat Y}$ is a function of the set of parameters $\{\mathbf{W}_l, \mathbf{R}_l\}$ at each layer. By optimizing the cross-entropy loss, we have the following problem formulation of node feature learning:
\begin{equation}
\min_{\{\mathbf{W}_l,\mathbf{R}_l\}} 
-\sum_{i=1}^N \sum_{c=1}^C \mathbf{Y}_{ic} \log(\hat{\mathbf{Y}}_{ic}(\mathbf{W}_l, \mathbf{R}_l)), 
\label{eq:opt_node}
\end{equation}
where $N$ is the number of input instances, $C$ is the number of classes, and $\mathbf{Y}$ denotes the ground truth label matrix.

\subsection{Problem Formulation of Joint Learning}

Integrating Eq. \eqref{eq:opt_graph_R} and Eq. \eqref{eq:opt_node}, we pose the joint learning of underlying graphs and node features as minimizing both GLR and the cross entropy.
The final problem formulation is 
\begin{equation}
\begin{split}
\min_{\{\mathbf{W}_l, \mathbf{R}_l\}} 
& -\sum_{i=1}^N \sum_{c=1}^C \mathbf{Y}_{ic} \log(\hat{\mathbf{Y}}_{ic}(\mathbf{W}_l, \mathbf{R}_l)) \\
& + \lambda \sum_l \sum_{\{i,j\}}\exp\left\{-\|\mathbf{R}^{\top}_l (\mathbf{f}^i_l - \mathbf{f}^j_l)\|_2^2 \right\} \, s_{i,j},
\end{split}
\label{eq:opt_R}
\end{equation}
where $\lambda > 0$ is a hyper-parameter for the trade-off between the cross entropy and the GLR term.

\vspace{-0.05in}
\section{Proposed Network Structure}
\label{sec:network}
Having discussed the problem formulation of joint learning, we elaborate on the proposed network architecture to realize dynamic learning of graphs and node features, with focus on the JLGCN Layer.  

\subsection{JLGCN Layer}

\begin{figure}[t]
\centering
\includegraphics[width=2.6in]{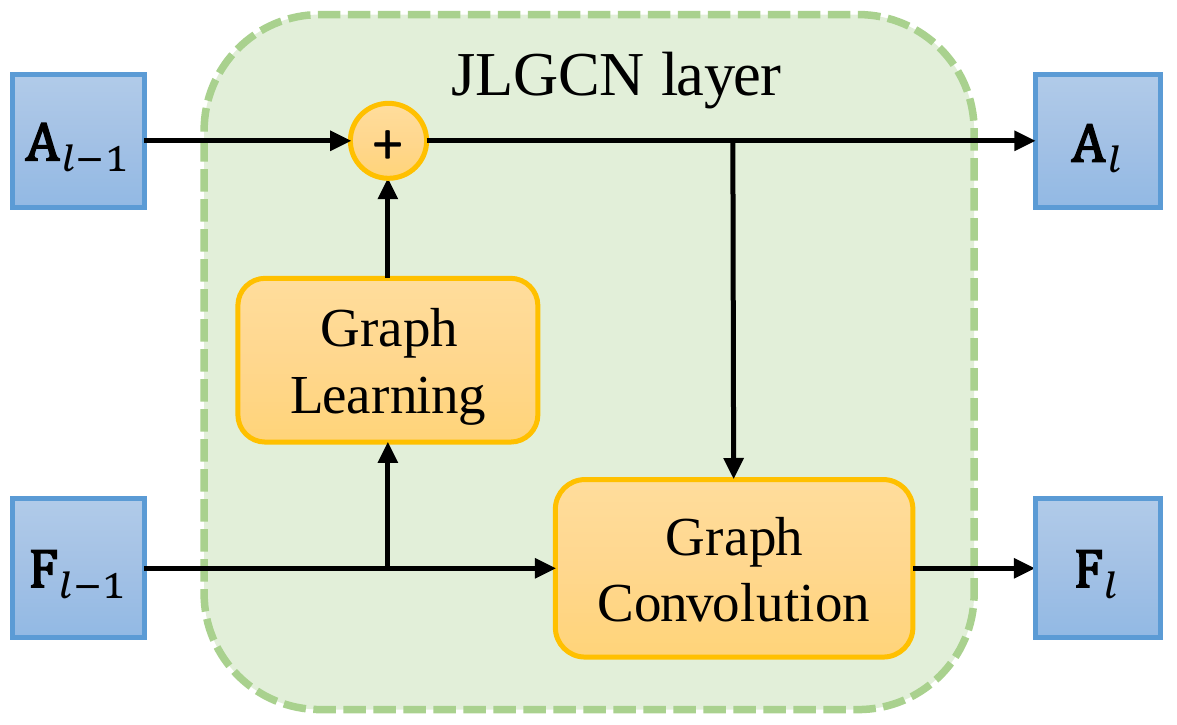}
\caption{Demonstration of a JLGCN layer. A graph is first learned from node features, and then employed to propagate node features via graph convolution.}
\label{fig:layer}
\end{figure}

As illustrated in Fig.~\ref{fig:layer}, the JLGCN layer is composed of two modules: the graph learning module and node feature learning module. 
The graph learning module learns a graph encoded by an $N \times N$ dense adjacency matrix $\mathbf{A}$ from input node features. 
$\mathbf{A}$ is then employed by the node feature learning module to propagate node features via our proposed graph convolution in Eq. \eqref{eq:gcn}. 
Both modules are jointly optimized according to the objective in Eq. \eqref{eq:opt_R}, \textit{i.e.}, the loss function is defined as 
\begin{equation}
    \begin{split}
E=& -\sum_{i=1}^N \sum_{c=1}^C \mathbf{Y}_{ic} \log(\hat{\mathbf{Y}}_{ic}(\{\mathbf{W}_l, \mathbf{R}_l\})) \\
& + \lambda \sum_l \sum_{\{i,j\}}\exp\left\{-\|\mathbf{R}^{\top}_l (\mathbf{f}^i_l - \mathbf{f}^j_l)\|_2^2 \right\} \, s_{i,j},
\end{split}
\label{eq:loss}
\end{equation}
The algorithmic details are provided in Alg. \ref{alg:layer}.

\begin{algorithm}[ht]
\caption{JLGCN layer} 
\label{alg:layer}
\hspace*{\algorithmicindent} \textbf{Input:} features $\mathbf{F}_{l-1}$ and adjacency matrix $\mathbf{A}_{l-1}$ from the previous layer \\
\hspace*{\algorithmicindent} \textbf{Output:} features $\mathbf{F}_l$ and learned adjacency matrix $\mathbf{A}_l$ \\
\hspace*{\algorithmicindent} \textbf{Initialize:} trainable parameters $\mathbf{R}_l, \mathbf{W}_l$
\begin{algorithmic}[1]
\State \textbf{Forward Pass:}
\State Compute distance metric $d_\mathbf{M}$ by Eq. \eqref{eq:m_distance_decompose}
\State Update $\mathbf{A}_l $ by Eq. \eqref{eq:gaussian_kernel} 
\State Update $\mathbf{F}_l $ by Eq. \eqref{eq:gcn}
\State \textbf{Backward Pass:}
\State Update $\mathbf{R}_l$ and $\mathbf{W}_l$ via backward propagation with the loss function defined in Eq. \eqref{eq:loss}. 
\end{algorithmic}
\end{algorithm}

\begin{figure*}[t]
\centering
\includegraphics[width=6in]{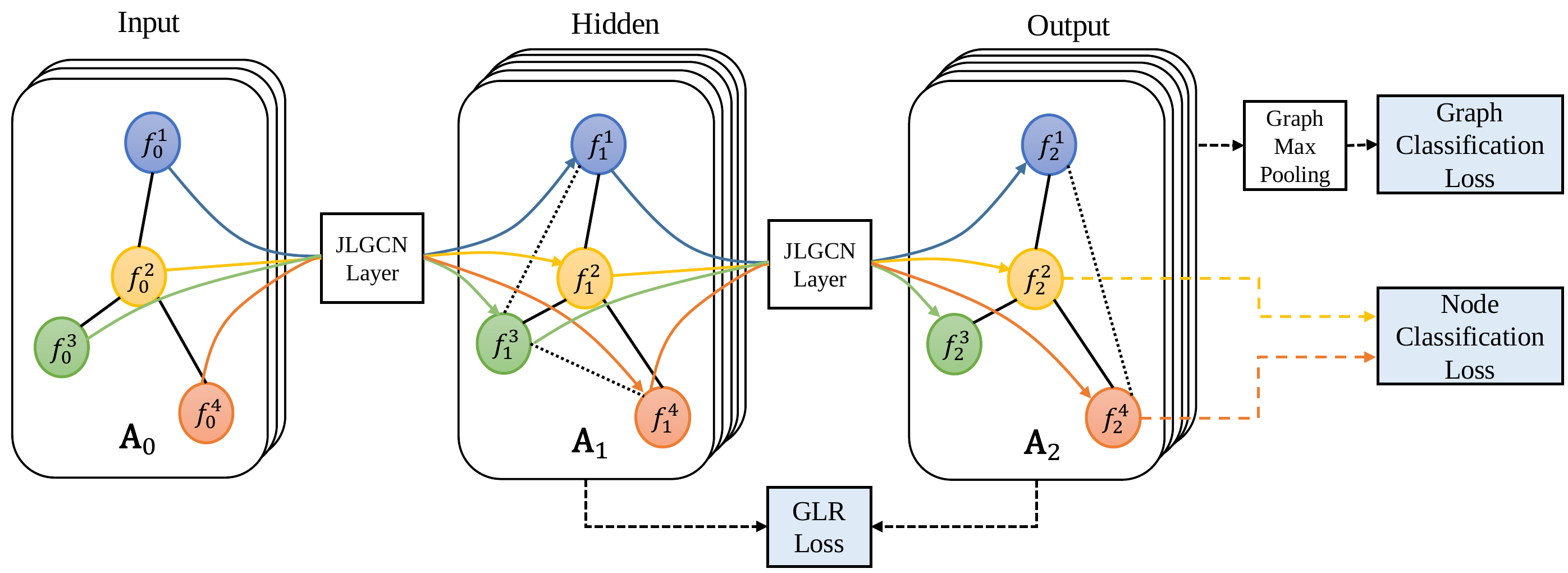}
\caption{The proposed JLGCN network architecture for node classification and graph classification tasks. The graph and node features at each layer are jointly updated.}
\label{fig:network}
\end{figure*}

\subsection{JLGCN Network}
In principle, our JLGCN layer can be integrated into any GCNN architecture for both node classification and graph classification tasks.
Fig.~\ref{fig:network} demonstrates the architecture of the JLGCN network.
Node features are extracted by alternatively learning an optimal graph based on features and propagating features based on the learned graph, which are then fed into the classification module for label prediction. 

\subsubsection{Node classification configuration}
We stack multiple JLGCN layers and deploy leaky ReLU activation \cite{xu2015empirical}. 
Dropout is applied to the last layer's input to reduce over-fitting. 
Features from the last layer is then employed for prediction of labels at each node.

\subsubsection{Graph classification configuration}
We stack multiple JLGCN layers followed by batch normalization \cite{ioffe2015batch} and leaky ReLU activation. Graph max pooling is applied to features from the last layer, which are then fed into a fully connected network for the classification of the entire graph.

Further, we perform feature concatenation at each layer $l$ with $\mathbf{f}_{l-1}$ from the previous layer for multi-level feature learning to enhance the model capacity. This leads to
\begin{equation}
\mathbf{f}_{l} = (\mathbf{f}_{l-1} || \mathbf{A}_l \mathbf{f}_{l-1}) \mathbf{W}_l,
\label{eq:gcn_concat}
\end{equation}
where $||$ is the concatenation operator in feature dimension. 
This essentially assigns more weighting to each node's own feature when no ground truth graph is available and accelerates the convergence.

\vspace{-0.05in}
\section{Experimental Results}
\label{sec:results}
In order to evaluate the performance of JLGCN, we apply it to the problems of semi-supervised node classification and point cloud classification, which are discussed in order as follows.

\subsection{Semi-supervised node classification}

\subsubsection{Datasets}
We test our method on three citation network datasets, \textit{i.e.}, Citeseer, Cora and Pubmed \cite{sen2008collective}. The statistics of the datasets are summarized in Tab. \ref{tab:citation_datasets}.

\begin{table}[ht]
\small
\centering
\begin{tabular}{p{0.4in}p{0.4in}p{0.3in}p{0.3in}p{0.4in}p{0.4in}}
\hline
\textbf{Dataset} & \textbf{Nodes} & \textbf{Edges} & \textbf{Classes} & \textbf{Features} & \textbf{Label Ratio} \\ \hline
Citeseer         & 3,327          & 4,732          & 6                & 3,703             & 0.036               \\
Cora             & 2,708          & 5,429          & 7                & 1,433             & 0.052               \\
Pubmed           & 19,717         & 44,338         & 3                & 500               & 0.003               \\ \hline
\end{tabular}
\caption{Summary of citation network datasets.} \label{tab:citation_datasets}
\end{table}

\subsubsection{Experimental settings}
We follow the experimental setup of previous work \cite{yang2016revisiting,Kipf2016SemiSupervisedCW,velivckovic2017graph}, and use the same data partition as in \cite{yang2016revisiting}. 
The features are $L2$-normalized before fed into the network.
Since our method builds a dense adjacency matrix among all nodes, we sub-sample 10,000 nodes for Pubmed dataset to evaluate our model due to memory limitation.
We set the number of JLGCN layers in our network to 2, and the number of hidden units in each layer to 16. For the dimension of the Mahalanobis distance metric matrix $\mathbf{R}\in \mathbb{R}^{K\times S}$, we set $S$ to 16 for both layers and initialize it randomly if not specified. We apply dropout with $p=0.5$ for both layer's input, and use leaky ReLU activation with negative slope $\alpha=0.2$ between two layers. We train our JLGCN for a maximum of 500 epochs using ADAM algorithm \cite{kingma2014adam} with a learning rate of 0.1 and weight decay of 0.0005. The learning rate is decayed by 0.5 every 100 epochs. The hyper parameter $\lambda$ for GLR is set to 0.0001. We report the average classification accuracy of 10 runs with different random seeds.

\subsubsection{Baselines}
We compare our JLGCN against the baseline of GCN \cite{Kipf2016SemiSupervisedCW} and Planetoid \cite{yang2016revisiting}, and also against some other graph neural network based semi-supervised learning methods, including Graph Attention Network (GAT) \cite{velivckovic2017graph} and Graph Markov Neural Network (GMNN) \cite{qu2019gmnn}. These methods use the same dataset partition as in our method for fair comparison with available codes.

\subsubsection{Results}

Tab.~\ref{tab:results_citation} shows the comparison results on three citation network datasets. The best results are marked in bold. 
Overall, we note that 
1) JLGCN outperforms the GCN baseline on all datasets significantly by only adding a graph learning module to each layer. This clearly demonstrates the effectiveness of jointly learning graphs and node features. Compared to the unweighted ground truth graph used in GCN, our learned graphs (as shown in Fig. \ref{fig:visual_cora}) enhance the semi-supervised learning results.
2) JLGCN also outperforms all the other recent work on the original Citeseer and Cora datasets and our sub-sampled Pubmed dataset. These results demonstrate the effectiveness of JLGCN in conducting semi-supervised classification tasks on graph data.

\begin{table}[ht]
\centering
\small
\begin{tabular}{p{1.4in}p{0.4in}p{0.4in}p{0.4in}}
\hline
\textbf{Algorithm} & \textbf{Citeseer} & \textbf{Cora} & \textbf{Pubmed*} \\ \hline
\begin{tabular}[c]{@{}l@{}}Planetoid\\\scriptsize\cite{yang2016revisiting}\end{tabular}     & 64.7  & 75.7  & 77.2   \\
\begin{tabular}[c]{@{}l@{}}GCN \\\scriptsize\cite{Kipf2016SemiSupervisedCW}\end{tabular}   & 70.3  & 81.5  & 78.4   \\
\begin{tabular}[c]{@{}l@{}}GAT\\\scriptsize\cite{velivckovic2017graph}\end{tabular}        & 72.5  & 83.0  & 78.8   \\
\begin{tabular}[c]{@{}l@{}}GMNN\\\scriptsize\cite{qu2019gmnn}\end{tabular}                  & 73.1  & 83.7  & 79.4   \\
\hline
JLGCN              & \textbf{73.7}            & \textbf{83.9}        & \textbf{79.7}             \\
\hline
\end{tabular}
\caption{Results of semi-supervised node classification. [*] means we sub-sample 10,000 nodes to perform the experiments.} \label{tab:results_citation} 
\end{table}

\subsubsection{Robustness test}

\begin{figure*}[t]
\centering
\subfloat[]{{\includegraphics[width=3in]{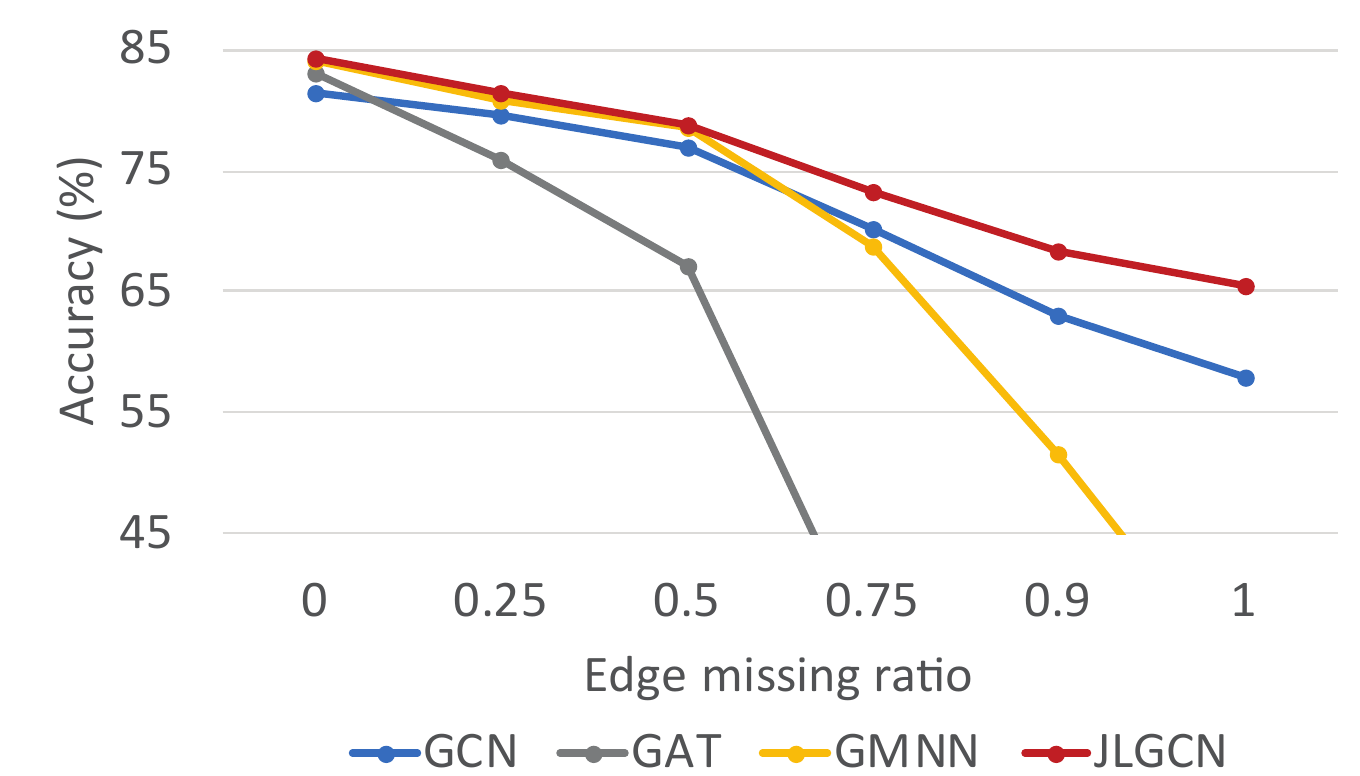} }}
\qquad
\subfloat[]{{\includegraphics[width=3in]{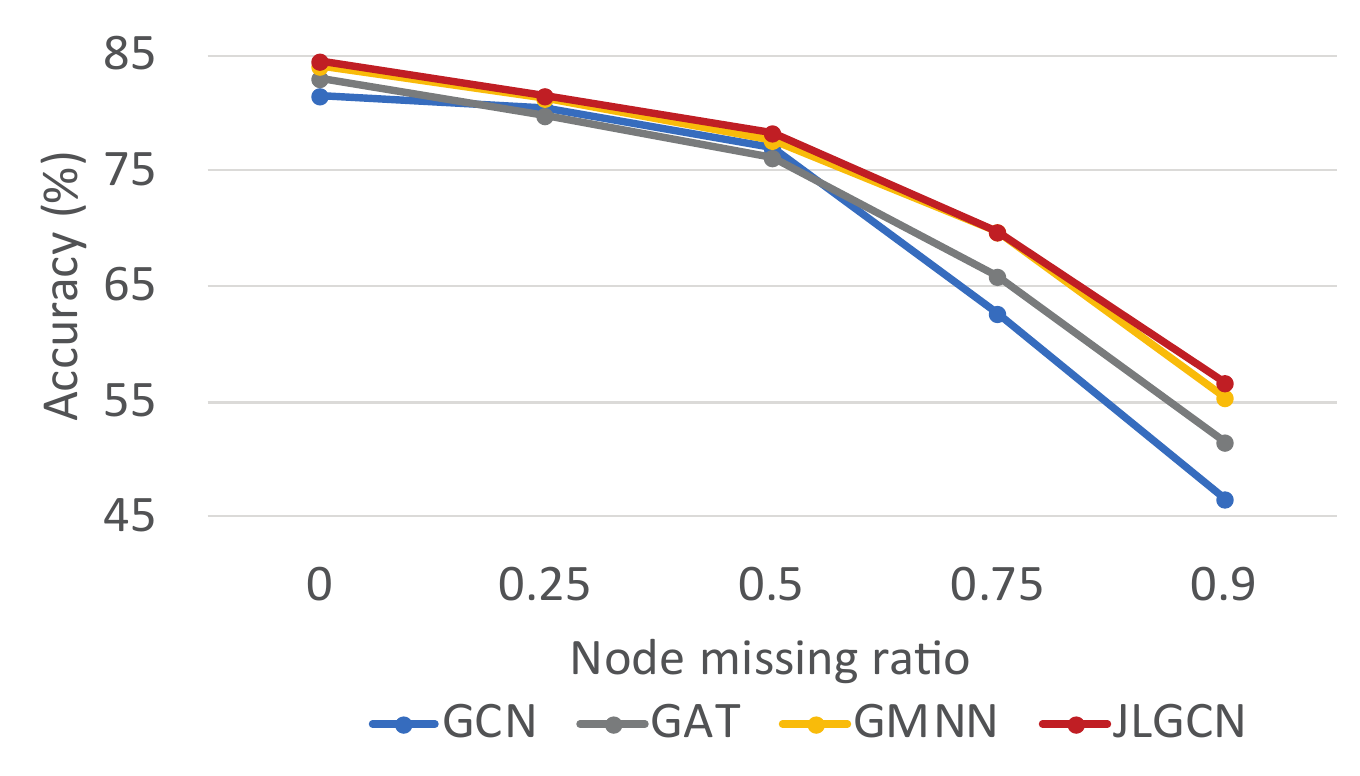} }}
\caption{Robustness test with different missing ratios of labeled nodes and edges on Cora dataset. (a) Accuracy with different edge missing ratios. (b) Accuracy with different node missing ratios.}
\label{fig:robust_citation}
\end{figure*}

We design two experiments to test the robustness of our model: 
1) we lower the label ratio by employing less training data. We adopt five missing ratios \{0, 0.25, 0.5, 0.75, 0.9\} of training nodes to train and evaluate;
2) we test the robustness of our model to incomplete ground truth graph, \textit{i.e.}, we randomly drop out edges in the ground truth graph, with edge missing ratios set as \{0, 0.25, 0.5, 0.75, 0.9, 1.0\}. 
When the edge missing ratio is 1.0, an empty ground truth graph is fed into the network. 
Since randomly initialized graph is hard to converge, we set $S=K$ for the transformation matrix $\mathbf{R}$ and initialize it as an identity matrix in this case. 
Fig. \ref{fig:robust_citation} shows our model significantly outperforms GCN and GAT, especially when the edge missing ratio is high.
This validates the superiority of our learned graph structure in terms of robustness. 

\subsubsection{Ablation study}
We first examine the effectiveness of the graph learning module on Cora dataset. 
Specifically, we compare with 
1) GCN, where the ground truth graph is employed; 
2) GCN + Euclidean, where we replace the proposed graph learning with manual graph construction using Euclidean distance metric to compute edge weights; 
3) JLGCN without dynamic updating of graphs, \textit{i.e.}, we only learn one graph from the input data feature.
As reported in Tab. \ref{tab:ablation}, our method leads to the best result when we adopt the Mahalanobis distance metric and learn different graphs at different layers jointly with node features from data.

Secondly, we evaluate the performance of JLGCN with different $\lambda$---the weighting parameter of GLR in the loss function---with respect to different edge missing ratios. 
As listed in Tab. \ref{tab:ablation}, we see that JLGCN achieves the highest accuracy when $\lambda$ is set to 0.0001. 
Also, GLR helps improve the performance especially when the ground truth graph is incomplete or even unavailable.

\begin{table}[ht]
\centering
\subfloat[]{
\begin{tabular}{ll}
\hline
\textbf{Algorithm} & \textbf{Accuracy}      \\ 
\hline
GCN (baseline)     & 81.5                   \\
GCN + Euclidean    & 83.1                   \\ 
JLGCN (single graph)  & 83.4                \\
\hline
JLGCN (multiple graphs) & \textbf{83.9}          \\ 
\hline
\end{tabular}}
\qquad
\subfloat[]{
\begin{tabular}{llll}
\hline
\multirow{2}{*}{$\mathbf{\lambda}$} & \multicolumn{3}{l}{\textbf{Edge Missing Ratios}} \\ \cline{2-4} 
                                 & 0              & 0.5            & 1.0            \\ \hline
0                                & 83.50           & 77.68          & 59.32          \\
0.0001                           & \textbf{83.88} & \textbf{77.90} & \textbf{60.52} \\
0.001                            & 82.76           & 76.96          & 59.84          \\
0.01                             & 82.45           & 76.46          & 59.40           \\ \hline
\end{tabular}}

\caption{Ablation study.(a) shows the results with different GLR weighting parameter $\lambda$ with respect to different edge missing ratios on Cora dataset. (b) shows the results of different graph learning methods on Cora dataset.}
\label{tab:ablation}
\end{table}

\subsubsection{Visualization of the learned graph}
\begin{figure}[t]
\centering
\subfloat[]{{\includegraphics[width=1.5in]{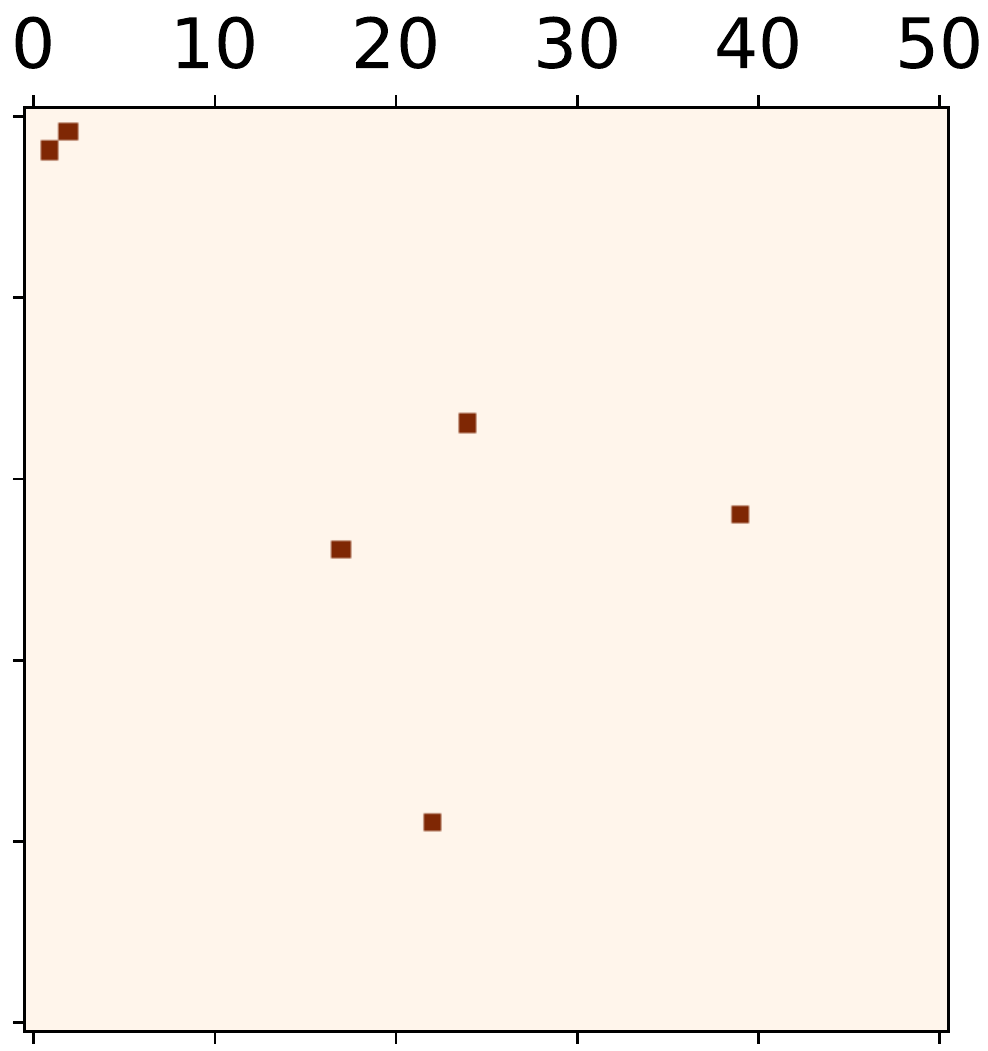} }}
\qquad
\subfloat[]{{\includegraphics[width=1.5in]{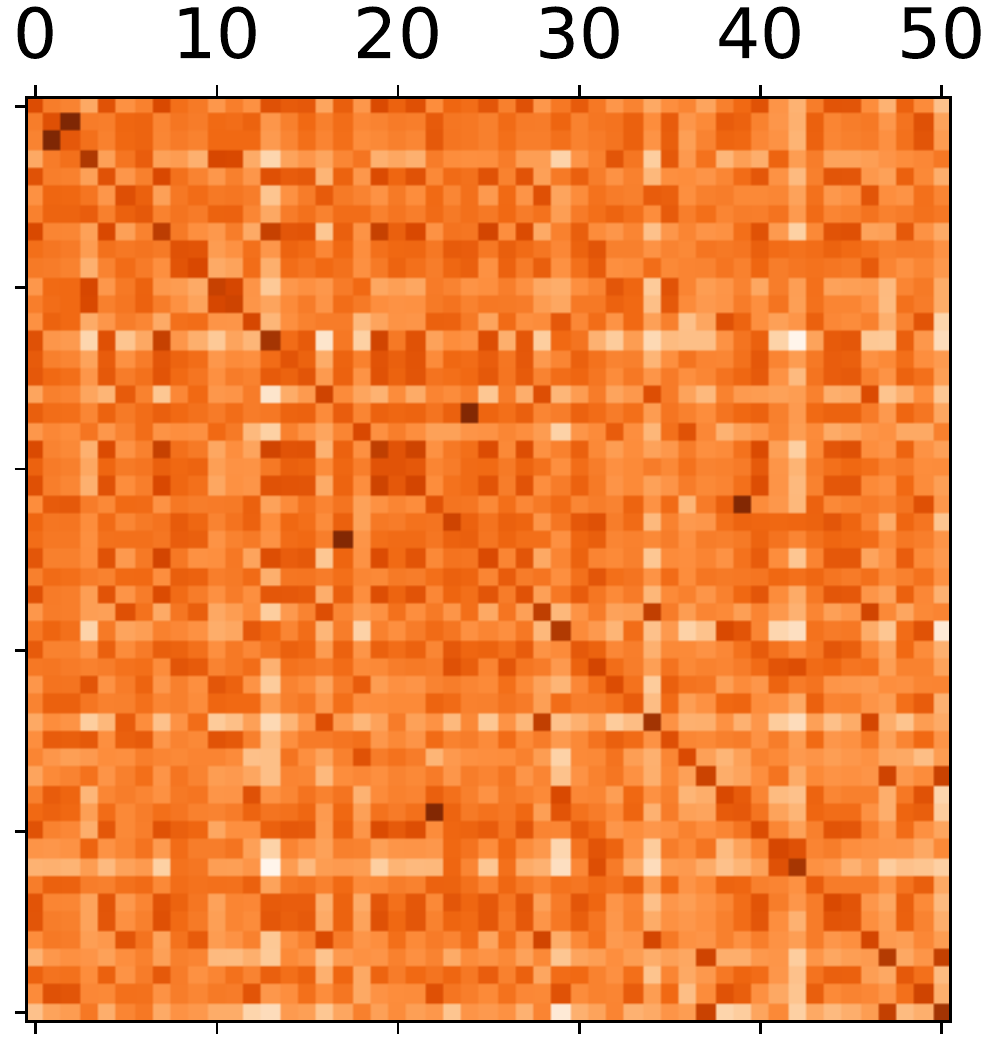} }}
\qquad
\subfloat[]{{\includegraphics[width=1.5in]{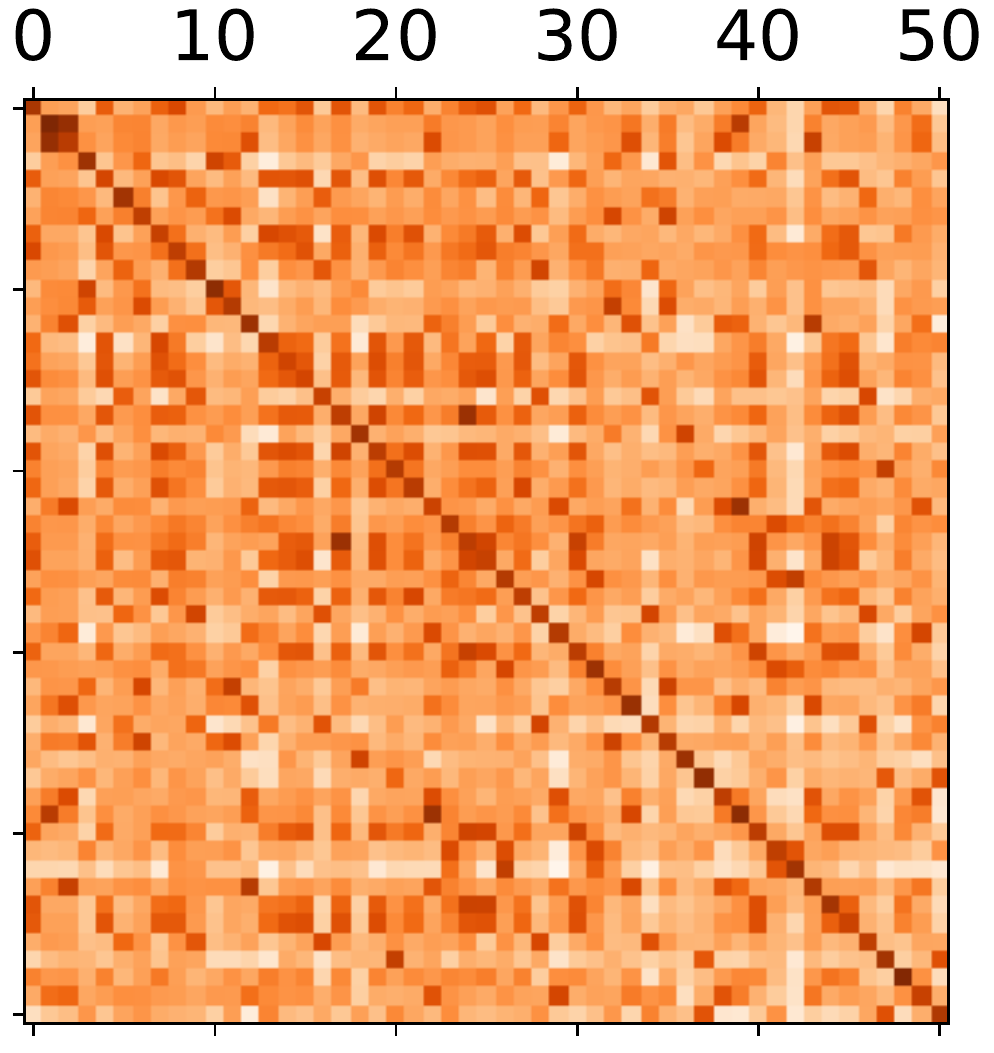} }}
\caption{Visualization of the learned graphs at different layers for the first 50 nodes on Cora dataset. (a) shows the ground truth graph, (b) is the learned graph for the first layer, and (c) for the second layer.}
\label{fig:visual_cora}
\end{figure}

We also visualize the learned graph at each layer in Fig. \ref{fig:visual_cora}, and compare them with the ground truth graph. The value of the matrix is log-transformed to make it more interpretable.
Note that the learned graph is both dense and weighted, which not only adds self-connection to the unweighted ground truth graph as the original GCN does, but also extracts additional hidden connections between similar nodes. 
Also, the learned graph at the second layer puts higher weighting to self-connection compared to the first layer, showing that different layers correspond to varying graphs as we assumed.

\subsection{Point cloud classification}
Furthermore, we test our model on point cloud classification to validate the effectiveness of our model when no ground truth graph is available.

\subsubsection{Datasets}
We test on ModelNet40 dataset for point cloud classification.
ModelNet40 includes 12,311 CAD models from 40 man-made categories, which are split into 9,843 for training and 2,468 for testing. 
As in previous work \cite{qi2017pointnet,qi2017pointnet++,wang2019dgcnn}, we uniformly sample 1024 points on mesh faces according to face area and normalize them into a unit sphere, where only coordinates of points are employed as input features.

\subsubsection{Experimental settings}
We stack 3 JLGCN layers to extract point cloud features as well as learn the underlying graph without any ground truth graph. 
The hidden units for each layer is \{64, 128, 1024\}. 
After extracting point cloud features, we deploys a permutation-invariant graph max-pooling to generate a global context feature for the entire point cloud, and employs a fully connected network with hidden units of \{512, 256, 40\} to classify the point cloud. 
The negative slope of leaky ReLU activation is set to 0.2 and dropout with $p=0.5$ is used in the fully connected classification network.
We train the network with a learning rate of 0.001 for 400 epochs and decay it by 0.5 every 40 epochs using ADAM. 
The batch size is set to 32, weight decay to 0.0001 and $\lambda$ to 0.01. 

\subsubsection{Baselines}
We also provide a GCN based model as the baseline, which empirically constructs a $k$-NN graph as the underlying graph. 
$k$ is set to 20 in the classification task based on the density of point clouds. 
We mainly compare our JLGCN model against the baseline GCN, and also compare against other state-of-the-art methods, including methods tailored for point cloud learning such as PointNet \cite{qi2017pointnet}, PointNet++ \cite{qi2017pointnet++}, PointGCN \cite{zhang2018graph}, RGCNN \cite{te2018rgcnn}, and DGCNN \cite{wang2019dgcnn}.

\subsubsection{Results}

Tab. \ref{tab:results_modelnet} presents results for ModelNet40 classification. 
Our method improves the baseline by about 2\% with only 3MB of parameters added, demonstrating the learned graph is superior to the empirical $k$-NN graph. 
Also, our method achieves comparable results compared to state-of-the-art methods for point loud learning, while our model is simpler and contains much smaller amount of parameters.
Note that, our method achieves the best performance among all spectral graph convolution based methods.

\begin{table}[ht]
\centering
\scriptsize
\begin{tabular}{p{0.3in}p{0.9in}p{0.4in}p{0.4in}p{0.4in}}
\hline
\textbf{Category} &
\textbf{Algorithm} & 
\textbf{\begin{tabular}[c]{@{}l@{}}Model Size\\(MB)\end{tabular}} &
\textbf{\begin{tabular}[c]{@{}l@{}}Mean Class\\Accuracy\end{tabular}} &
\textbf{\begin{tabular}[c]{@{}l@{}}Overall\\Accuracy\end{tabular}} \\ 
\hline
\begin{tabular}[c]{@{}l@{}}Pointwise\\MLP\end{tabular} & \begin{tabular}[c]{@{}l@{}}PointNet\\\cite{qi2017pointnet}\end{tabular} & 40   & 86.0                        & 89.2                      \\
\begin{tabular}[c]{@{}l@{}}Pointwise\\MLP\end{tabular} & \begin{tabular}[c]{@{}l@{}}PointNet++\\\cite{qi2017pointnet++}\end{tabular}    & 12   & -                           & 90.7                      \\
\begin{tabular}[c]{@{}l@{}}Spectral\\GCNN\end{tabular} & \begin{tabular}[c]{@{}l@{}}PointGCN\\\cite{zhang2018graph}\end{tabular}        & 41    & 86.1                        & 89.5                      \\
\begin{tabular}[c]{@{}l@{}}Spectral\\GCNN\end{tabular} & \begin{tabular}[c]{@{}l@{}}RGCNN\\\cite{te2018rgcnn}\end{tabular}             & 22   & 87.3                        & 90.5                      \\
\begin{tabular}[c]{@{}l@{}}Spatial\\GCNN\end{tabular} & \begin{tabular}[c]{@{}l@{}}DGCNN\\\cite{wang2019dgcnn}\end{tabular}        & 21   & \textbf{90.2}               & \textbf{92.9}             \\ 
\hline
\begin{tabular}[c]{@{}l@{}}Spectral\\GCNN\end{tabular} & GCN (baseline)                      & \textbf{10}   & 84.2                        & 88.7                      \\ 
\begin{tabular}[c]{@{}l@{}}Spectral\\GCNN\end{tabular} & JLGCN                                & 13   & 87.2                        & 90.8                      \\ 
\hline
\end{tabular}
\caption{Results of point cloud classification on ModelNet40.} \label{tab:results_modelnet} 
\end{table}

\subsubsection{Robustness test}
\begin{figure}[ht]
\centering
\includegraphics[width=3in]{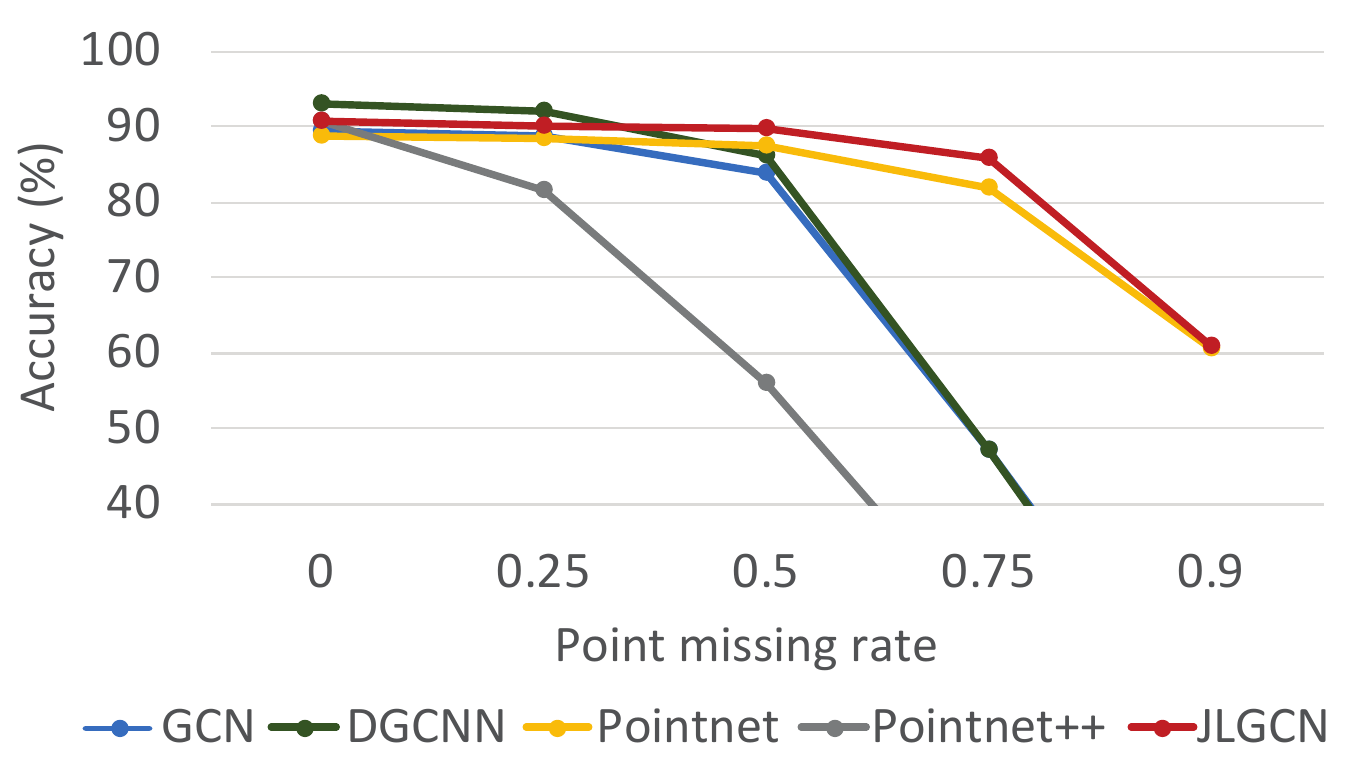}
\caption{Robustness test with different missing ratios of point clouds.}
\label{fig:robust_pc}
\end{figure}

Further, we test the robustness of our model when the point cloud is of low density. 
We randomly drop out points with missing ratios \{0, 0.25, 0.5, 0.75, 0.9\}. 
As shown in Fig. \ref{fig:robust_pc}, our model outperforms the GCN baseline significantly, and keeps high accuracy even when the point cloud density is quite low.

\vspace{-0.05in}
\section{Conclusion}
\label{sec:conclude}
We propose joint learning of graphs and node features in Graph Convolutional Neural Networks, which dynamically learns graphs adaptive to the structure of node features in different layers. 
In particular, we optimize an underlying graph kernel via distance metric learning with the Mahalanobis distance employed. 
The metric matrix is decomposed into a low-dimensional matrix and optimized based on Graph Laplacian Regularizer. 
Extensive experiments demonstrate the superiority and robustness of our method in semi-supervised learning and point cloud learning problems. 

\bibliographystyle{unsrt}

\end{document}